\documentclass{article}

\usepackage[final]{neurips_2024}

\usepackage{amsmath,amsfonts,bm}

\def\eqref#1{equation~\ref{#1}}

\def\1{\bm{1}}

\DeclareMathAlphabet{\mathsfit}{\encodingdefault}{\sfdefault}{m}{sl}
\SetMathAlphabet{\mathsfit}{bold}{\encodingdefault}{\sfdefault}{bx}{n}

\usepackage[utf8]{inputenc} 
\usepackage[T1]{fontenc}    
\usepackage{hyperref}       
\usepackage{url}            
\usepackage{booktabs}       
\usepackage{amsfonts}       
\usepackage{nicefrac}       
\usepackage{microtype}      
\usepackage{xcolor}         
\usepackage{multirow}
\usepackage{adjustbox}
\usepackage{amsmath}
\usepackage{pifont}
\usepackage{colortbl}
\usepackage{wrapfig}
\usepackage{makecell}
\usepackage{caption}
\usepackage{subcaption}
\usepackage{graphicx}
\usepackage{mathrsfs}
\usepackage{amssymb}
\usepackage{mathtools}
\usepackage{amsthm}
\usepackage{booktabs}
\usepackage{multirow}
\usepackage{listings}
\usepackage{xcolor}
\usepackage{algorithm}
\usepackage{algpseudocode}
\usepackage{graphicx}
\usepackage{float}
\usepackage{fancyvrb}
\usepackage{mdframed} 
\usepackage{verbatim}
\usepackage{tcolorbox}
\tcbuselibrary{listings}
\usepackage{tcolorbox}
\usepackage{color}
\tcbuselibrary{listings}
\usepackage{geometry}
\usepackage{enumitem}
\usepackage{calc}
\usepackage{tcolorbox}
\usepackage{textcomp} 
\tcbuselibrary{listings, breakable}
\usepackage{spverbatim}
\usepackage{array}
\usepackage{xspace}    
\newcommand\eg {{\it e.g., }}

\definecolor{lightergray}{gray}{0.05}
\definecolor{boxgray}{rgb}{.94, .94, .94}
\newtcolorbox{mybox}[1][]{
  colback=white!95!gray, 
  colframe=gray!75!black, 
  fonttitle=\bfseries\large\ttfamily, 
  title=#1,
  coltitle=white, 
  colbacktitle=gray!75!black, 
  sharp corners,
  boxrule=1pt,
  left=4pt,
  right=4pt,
  top=4pt,
  bottom=4pt,
  breakable,
  width=\dimexpr\textwidth+50pt\relax, 
  enlarge left by=-1pt, 
  enlarge right by=-1pt, 
}
\definecolor{codebg}{rgb}{0.95,0.95,0.95}
\title{Distributionally robust self-supervised learning for tabular data}

\author{%
  Shantanu Ghosh\thanks{Work completed during an internship at Amazon} \\
  Boston University \\
  \texttt{shawn24@bu.com} \\
  \And
  Tiankang Xie \\
 Amazon \\
  \texttt{xietiank@amazon.com} \\
  \And
  Mikhail Kuznetsov \\
  Amazon \\
  \texttt{mikuzne@amazon.com} \\
}

\begin{document}

\maketitle

\begin{abstract}
\label{sec:abstract}
Machine learning (ML) models trained using Empirical Risk Minimization (ERM) often exhibit systematic errors on specific subpopulations of tabular data, known as error slices. Learning robust representation in the presence of error slices is challenging, especially in self-supervised settings during the feature reconstruction phase, due to high cardinality features and the complexity of constructing error sets. Traditional robust representation learning methods are largely focused on improving worst group performance in supervised settings in computer vision, leaving a gap in approaches tailored for tabular data. We address this gap by developing a framework to learn robust representation in tabular data during self-supervised pre-training. Our approach utilizes an encoder-decoder model trained with Masked Language Modeling (MLM) loss to learn robust latent representations. This paper applies the Just Train Twice (JTT) and Deep Feature Reweighting (DFR) methods during the pre-training phase for tabular data. These methods fine-tune the ERM pre-trained model by up-weighting error-prone samples or creating balanced datasets for specific categorical features. This results in specialized models for each feature, which are then used in an ensemble approach to enhance downstream classification performance. This methodology improves robustness across slices, thus enhancing overall generalization performance. Extensive experiments across various datasets demonstrate the efficacy of our approach. The code is available: \url{https://github.com/amazon-science/distributionally-robust-self-supervised-learning-for-tabular-data}.
\end{abstract}

\section{Introduction}
\label{sec:intro}
ERM-trained ML models often exhibit systematic errors on specific subpopulations of data, known as error slices. Supervised robust representation learning techniques \eg DFR~\citep{kirichenko2022last}, GroupDRO~\citep{Sagawa2020}, JTT~\citep{Liu2021} mitigate the model's error rate on worst group subpopulation. While researchers develop such mitigation methods traditionally in computer vision for supervised setup, there is a lack of approaches to learn robust representation during self-supervised pre-training for tabular data. 
This paper addresses this gap by developing error slices during the self-supervised learning phase, focusing on data reconstruction to learn robust representations across features. These representations enhance overall the downstream classification performance, leading to better generalization across various data subpopulations other than the worst group.


The literature on robust representation learning is extensive. GroupDRO~\citep{Sagawa2020}  improves model generalization by minimizing the worst-case loss over predefined groups within the dataset. This approach uses regularization techniques to ensure consistent performance of a model across all groups, mitigating the risk of poor performance on underrepresented subpopulations.  Just Train Twice (JTT)~\citep{Liu2021} introduces a two-stage training process to enhance robustness without explicit group labels. In the first stage, JTT identifies ``difficult'' examples by training an ERM-based model and identifying samples where the model misclassifies. In the second stage, the model is retrained by up-weighting these difficult examples to improve overall robustness. This up-weighting strategy allows JTT to enhance model performance across diverse data subpopulations.  Also, JTT enhances model robustness without requiring prior knowledge of specific error-prone groups. Recently, DFR~\citep{kirichenko2022last} demonstrated that retraining only the last layer of a neural network using group-balanced validation data.  However, these approaches are primarily tailored for computer vision, with no established robustness methods for tabular data. Applying robust training strategy \eg JTT or DFR in supervised label prediction settings for tabular data involves discovering, up-weighting error sets or creating balanced datasets, which is relatively straightforward. Also, for supervised training, these methods focus on improving worst-group performance. However, applying them during the self-supervised reconstruction phase is challenging due to the high cardinality features and the complexity of constructing error sets for multiple features. Despite these challenges, employing them during the reconstruction phase is crucial for learning robust representations. This work addresses these challenges by adapting JTT and DFR to the reconstruction phase, enhancing overall downstream classification performance across various error slices.

Our contribution: In this paper, we propose two novel strategies using JTT and DFR for learning robust representations for tabular data. Unlike traditional supervised setups for label prediction, our approach employs a self-supervised strategy during the reconstruction phase. Each strategy has two stages -- ERM pre-training and robust representation learning. The ERM pre-training stage employs an encoder-decoder model using Masked Language Modeling (MLM) loss to learn latent representations, which are then used for downstream classification tasks. The JTT-based robust representation learning first creates an error set to identify the hard samples. Specifically, for each categorical feature, it identifies samples where the reconstruction fails, to create an error set. The DFR-based method employs DFR to create a balanced dataset for each category from the validation set. Next, we finetune the encoder-decoder model either by up-weighting these samples (strategy 1, JTT) or using the balanced dataset (strategy 2, DFR). This finetuning stage focuses on the specific decoder head for that feature, while keeping others fixed. This process results in a specialized model for each categorical feature. During inference, we employ an ensemble approach: we estimate the reconstruction loss for all features for a given sample and identify the feature with the maximum loss. We then select the representation from the specialized model corresponding to this feature, ensuring the use of a robust representation for classification. This approach utilizes the specialized capabilities of each model per categorical feature to enhance classification accuracy.
Finally, we fine-tune the classifier using these robust representations and estimate performance across different feature categories. This approach in the reconstruction phase, enhances generalization and achieves superior performance across diverse data subpopulations compared to standard ERM models. Extensive experiments across various datasets and architectures validate our method.

\section{Method}
\label{sec:method}
\begin{figure}[h!]
    \centering
    \includegraphics[width=\textwidth]{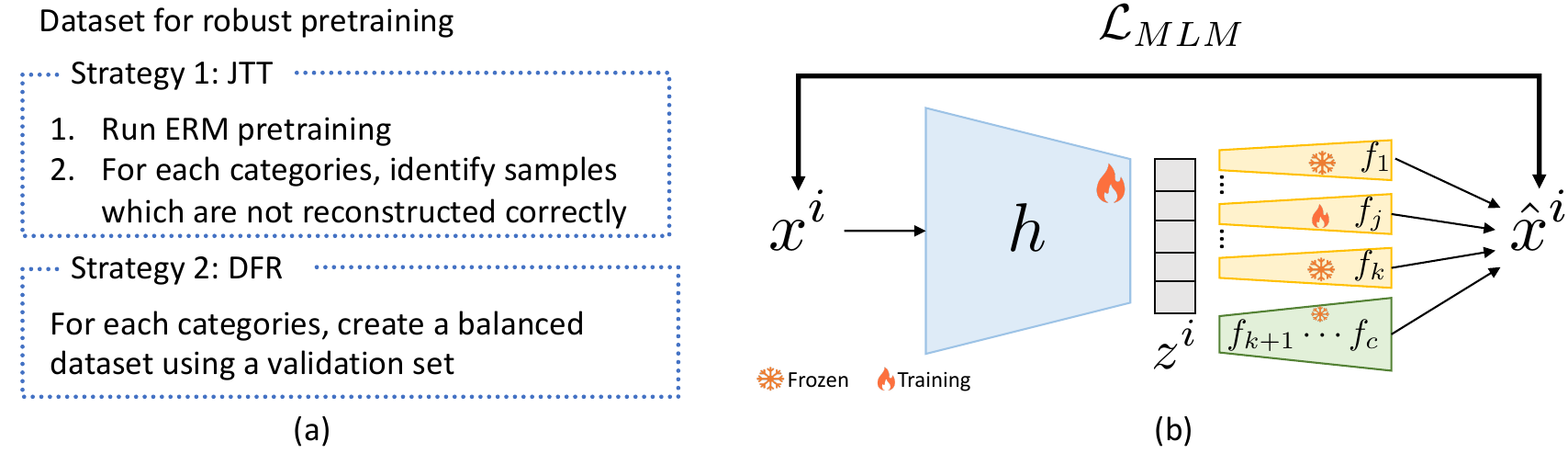}
    \caption{Schematic of our method. (a) Dataset construction of robust pretraining using JTT and DFR. (b) Robust pre-training strategy using MLM loss. We train the encoder $h$ and the reconstruction head of the $j^{th}$ feature $f_j$ for each sample. The embedding $z$ will be used for downstream tasks. We do this for all categorical features, obtaining a pre-trained model per category. 
    }
    \label{fig:schematic}
\end{figure}

Our pretraining strategy consists of two stages: Stage 1 ERM pre-training and Stage 2 robust representation learning. The training process aims to build a robust model capable of both feature reconstruction and target prediction accurately. Stage 1 optimizes Masked Language Modeling (MLM) loss for feature reconstruction using ERM~\citep{Vapnik1999} to learn latent representations. Motivated by the expert-based model~\citep{pmlr-v202-ghosh23c, ghosh2023distilling} and the ensemble bias-specific mitigation method discussed in~\cite{ghosh2024ladderlanguagedrivenslice}, stage 2 employs two independent strategies using JTT and DFR to learn robust representation. Strategy 1, using JTT, identifies the samples that are not reconstructed correctly, forming the error set for each categorical feature. Next, for each category, it upweights the samples in the error set, learning specific models per category. For strategy 2, using DFR-based pre-training for each category, phase 2 constructs a balanced validation dataset and learns models for each category. For the downstream classification, we employ an ensemble approach to construct the representation. Fig.~\ref{fig:schematic} depicts our method.  Lastly, a classifier is trained on these representations to predict target labels. Algorithms~\ref{algo: JTT} and ~\ref{algo: DFR} present our proposed algorithm.

\subsubsection{Notation:}
The dataset $\{\mathcal{X}, \mathcal{Y}\}$ consists of input features $\boldsymbol{x} \in \mathcal{X}$ and target labels $\mathcal{Y}$. The features $\mathcal{X}$ include $k$ categorical features $(\boldsymbol{x}_1, \boldsymbol{x}_2, \cdots, \boldsymbol{x}_k)$ and $c$ continuous features $(x_{k+1},x_{k+2} \cdots x_{c})$. Also, $N$ denotes the number of samples.
The model architecture includes an encoder-decoder framework.
$h$ denotes the encoder of the model, mapping the input $x$ to a latent representation $h(x)$.
For $k$ categorical features, $f_1, f_2, \cdots, f_k$ represents the decoder heads reconstructing the categorical features. Similarly, for $c$ continuous features, $f_{k+1}, f_{k+2}, \cdots, f_{k+c}$ represent the decoder heads reconstructing the continuous features.

\subsubsection{Stage1: ERM with MLM Loss }
In Phase 1A, the model is trained using a MLM loss, which aims to reconstruct the input features from the latent representation $h(x)$. The MLM loss, $\mathcal{L}_{\text{MLM}}$, is calculated as follows: 

\begin{align}
    \label{Eq.MLM}
    \mathcal{L}_{\text{MLM}} = \frac{1}{N} \sum_{i=1}^{N} \left( \sum_{j=1}^{k} \mathcal{L}_{\text{cat}}(f_j(h(\boldsymbol{x}^i)), x_j^i) + \sum_{l=1}^{c} \mathcal{L}_{\text{cont}}(f_{k+l}(h(\boldsymbol{x}^i)), x_{k+l}^i) \right) ,
\end{align}

where $f_j(h(\boldsymbol{x}^i)$ is the output of the $j^{th}$ categorical head of the decoder for sample $i$, $x_j^i$ is the true value of the $j^{th}$ categorical head of the decoder for sample $i$, $\mathcal{L}_{\text{cat}}$ is the categorical reconstruction loss (cross-entropy), $f_{k+l}(h(\boldsymbol{x}_i))$ is the output of the $l^{th}$ continuous head of the decoder for sample $i$, $x_{k+l}^i$ is the true value of the $l^{th}$ continuous feature for sample $i$, $\mathcal{L}_{\text{cont}}$ is the continuous reconstruction loss (mean squared error).

\subsubsection{Stage2: Robust representation learning for tabular data}
\textbf{Strategy 1: JTT-based pre-training.}
This strategy focuses on improving the robustness of $h(x)$ by employing the JTT method. For each categorical feature $j$, we define an error set consisting of samples where the predicted categorical feature value does not match the true value, i.e., $x_j^i \neq \hat{x}_j^i$ where $\hat{x}_j^i = f_j(h(\boldsymbol{x}^i))$. We weigh these samples, resulting in a new loss function $\mathcal{L}_{\text{MLM}}^j$ for training the model specialized for feature $j$:

\begin{align}
    \tilde{\mathcal{L}}_{\text{MLM}} = \frac{1}{N} \sum_{i=1}^{N} \left(w_j^i \mathcal{L}_{\text{cat}}(f_j(h(x^i)), x_j^i) + \bigg(\sum_{\substack{m=1, \\ m\neq j}}^{k} \mathcal{L}_{\text{cat}}(f_m(h(x^i)), x_m^i) + \sum_{l=1}^{c} \mathcal{L}_{\text{cont}}(f_{k+l}(h(x^i)), x_{k+l}^i) \bigg)\right)
\end{align}

where $w_j^i$ is the upweight factor for sample $i$ concerning feature $j$, typically a hyperparameter to tune. Note that, in this phase, we only train h, the encoder and $f_j(.)$, the $j^{th}$ head of the deocder of the model keeping the heads of other categorical features and continuous features fixed. This training strategy debias the model for the specific features.

\textbf{Strategy 2: DFR-based pre-training.} 
This strategy focuses on enhancing the robustness of $h(\boldsymbol{x})$ through the DFR method. For each categorical feature $j$, we create a balanced validation set $\mathcal{D}_{bal}^j$ by selecting samples that represent different feature categories proportionally. This balanced set ensures that the model learns representations that are robust across diverse subpopulations. Finally for for feature $j$, we optimize the loss in Eq.~\ref{Eq.MLM} using $\mathcal{D}_{bal}^j$.
In contrast to JTT, DFR focuses on training the model using a balanced dataset for each categorical feature $j$, ensuring that the learned representations are less biased and more generalizable across different feature categories. Unlike the JTT phase, in this phase, we only train $f_j(.)$, the $j^{th}$ head of the decoder, while keeping the heads of other features fixed. Thus it ensures robustness to variations within each feature category, improving overall generalization.

\textbf{Downstream classifier training with feature-specific model selection}
Motivated by the ensemble mitigation approach~\cite{ghosh2024ladderlanguagedrivenslice}, for each test sample $x^*$, we calculate the reconstruction losses for each specialized model corresponding to the categorical features, determining the feature $j^*$ with the maximum loss:
\begin{align}
    j^* = \arg\max_j \mathcal{L}_{\text{cat}}(f_j(h(\boldsymbol{x}^*)), x_j^*)
\end{align}
The representation from the model $j^*$ is then used to train the classifier $g$, with the supervised loss:
\begin{align}
    \mathcal{L}_{\text{sup}} = \frac{1}{N} \sum_{i=1}^{N} \text{CrossEntropy}(g(h_{j^*}(\boldsymbol{x}_i)), Y^i)
\end{align}

\section{Experiments}
\label{sec:experiments}
We conduct using the bank~\cite{bank_marketing_222} and census~\cite{census_income_20} datasets. Refer to Appendix~\ref{sec: datasets} and Appendix~\ref{sec: experiments} for dataset and experimental details.

\begin{figure}[h!]
    \centering
    \includegraphics[width=\textwidth]{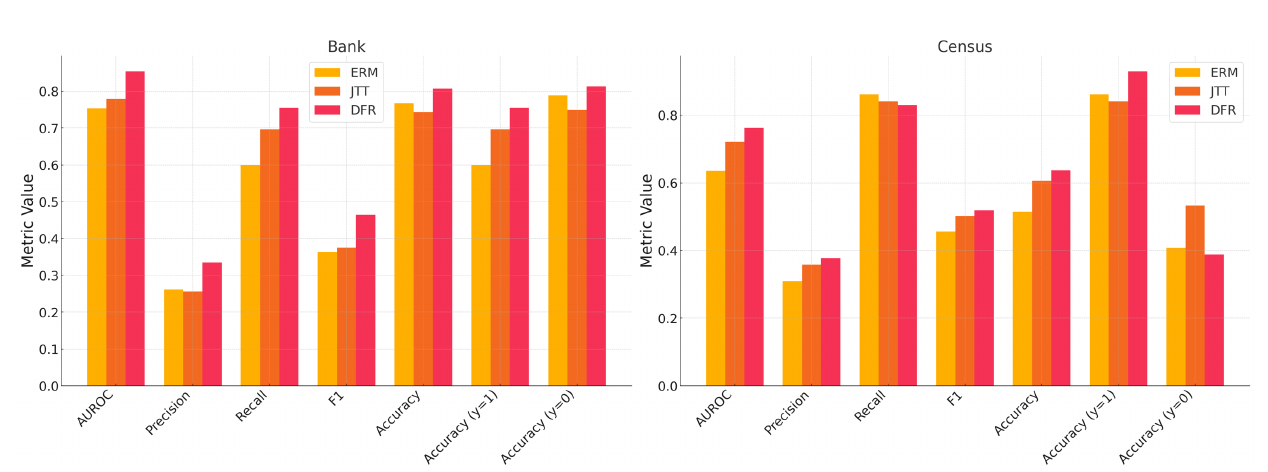}
    \caption{Comparing overall performance of the downstream classifiers using ERM, JTT and DFR}
    \label{fig:dfr_jtt_bank}
\end{figure}



\label{sec:results}

\section{Results}
\label{sec:results}
Fig. \ref{fig:dfr_jtt_bank} (left) and Fig. \ref{fig:dfr_jtt_bank} (right) illustrate the performance comparison between DFR and JTT on the Bank and Census datasets, respectively. DFR consistently outperforms JTT and ERM across these metrics, demonstrating its effectiveness in learning robust representations through balanced validation sets. DFR  creates a balanced dataset for each feature ensuring that the model is not overly influenced by majority classes or features. This leads to a more generalized learning process, allowing DFR to better capture the underlying structure of the data, especially in scenarios where certain features are prone to bias or imbalance. The result is a higher AUROC (14\% and 25\% gains over ERM for Bank and Census datasets, respectively).
\label{sec:results}


\section{Conclusion and limitation}
\label{sec:conclusion}
Our approach significantly improves model robustness and generalization across diverse subpopulations. Extensive experiments validate the effectiveness of our method, making it a promising solution for enhancing fairness and accuracy in tabular model training. A key future goal is to extend our framework to handle more complex tabular datasets with high cardinality features. Moreover, we aim to integrate causal inference techniques to further mitigate underlying biases(~\cite{ghosh2021propensity, ghosh2021deep, ghosh2023dr, prosperi2021causal}).
\label{sec:conclusion}

\bibliography{iclr2025_conference}
\bibliographystyle{plainnat}

\appendix
\section{Appendix}
\section{Related Work}
\subsection{Slice discovery} 
In computer vision, researchers have developed slice discovery frameworks to identify subpopulations where models consistently make errors that follow specific patterns. Early works on unstructured data include Spotlight, Multiaccuracy, and FailureModeAnalysis. They typically project data into a representation space and identify error slices using clustering or dimensionality reduction techniques. They are primarily assessed through limited slice configurations or qualitative analysis. Recent works include using the vision language representation space for slice discovery and obtaining SOTA results. These methods include the following:

\textbf{DOMINO}: DOMINO~\citep{eyuboglu2022domino} identifies systematic errors in machine learning models by leveraging cross-modal embeddings. It operates in three main steps: embedding, slicing, and describing.

1. Embedding: Domino uses cross-modal models (\eg CLIP) to embed inputs and text in the same latent space. This enables the incorporation of semantic meaning from text into input embeddings, which is crucial for identifying coherent slices.

2. Slicing: It employs an error-aware mixture model to detect underperforming regions within the embedding space. This model clusters the data based on embeddings, class labels, and model predictions to pinpoint areas where the model performance is subpar. The mixture model ensures that identified slices are coherent and relevant to model errors.

3. Describing: Domino generates natural language descriptions for the discovered slices. It creates prototype embeddings for each slice and matches them with text embeddings to describe the common characteristics of the slice. This step provides interpretable insights into why the model fails on these slices.

\textbf{FACTS}: FACTS~\citep{yenamandra2023facts} (First Amplify Correlations and Then Slice) aims to identify bias-conflicting slices in datasets through a two-stage process:

1. Amplify Correlations: This stage involves training a model with a high regularization term to amplify its reliance on spurious correlations present in the dataset. This step helps segregate biased-aligned from bias-conflicting samples by making the model fit a simpler, biased-aligned hypothesis.
2. Correlation-aware Slicing: In this stage, FACTS uses clustering techniques on the bias-amplified feature space to discover bias-conflicting slices. The method identifies subgroups where the spurious correlations do not hold, highlighting areas where the model underperforms due to these biases.

FACTS leverages a combination of bias amplification and clustering to reveal underperforming data slices, providing a foundation for understanding and mitigating systematic biases in machine learning models.

Assuming that an existing feature causes the error slice, identifying error slices in tabular data is straightforward. For tabular data, we can enumerate different feature categories within a particular class. This allows us to determine which specific feature category has a higher error rate than the overall rate. For example, in the UCI-Bank dataset, we aim to determine if an error slice exists for the feature `job'. We do this by estimating the error rate for each job category within a specific target class and comparing it to the overall error rate for that class. We identify an error slice in any job category that has a significantly higher error rate than the average error rate for the given target class.

\subsection{Error mitigation}
Error mitigation aims to improve the subgroup's performance where the model performs worst. These algorithms and provide detailed descriptions for each category below:

\textbf{Vanilla}: The empirical risk minimization(ERM)~\citep{Vapnik1999} algorithm seeks to minimize the cumulative error across all samples.

\textbf{Subgroup Robust Methods}: GroupDRO~\citep{Sagawa2020} proposes a robust optimization strategy, which enhances ERM by prioritizing groups with higher error rates. CVaRDRO~\citep{Duchi2018} is a variant of GroupDRO that dynamically assigns weights to data samples with the highest losses. LfF~\citep{Nam2020} concurrently trains two models: the first is biased, and the second is de-biased by re-weighting the loss gradient. Just Train twice (JTT)~\citep{Liu2021}  proposes an approach that initially trains an ERM model to identify minority groups in the training set, followed by a second ERM model where the identified samples are re-weighted. 
LISA~\citep{Yao2022} utilizes invariant predictors through data interpolation within and across attributes. DFR~\citep{kirichenko2022last} suggests first training an ERM model and then retraining the final layer using a balanced validation set with group annotations.

\textbf{Data Augmentation}: Mixup~\citep{Zhang2018} proposes an approach that performs ERM on linear interpolations of randomly sampled training examples and their corresponding labels.

\textbf{Domain-Invariant Representation Learning}: Invariant Risk Minimization~\citep{Arjovsky2019} (IRM)learns a feature representation such that the optimal linear classifier on this representation is consistent across different domains. MMD~\citep{Li2018} utilizes maximum mean discrepancy~\citep{Gretton2012} to match feature distributions across domains. Note that all methods in this category necessitate group annotations during training.

\textbf{Imbalanced Learning}: Focal~\citep{Lin2017} introduces Focal Loss, which reduces the loss for well-classified samples and emphasizes difficult samples. CBLoss~\citep{Cui2019} suggests re-weighting by the inverse effective number of samples. LDAM~\citep{Cao2019} employs a modified margin loss that preferentially weights minority samples. CRT and ReWeightCRT~\citep{Kang2020} (a re-weighted variant of CRT) decompose representation learning and classifier training into two distinct stages, re-weighting the classifier using class-balanced sampling during the second stage.

\textbf{Using Large language models} LADDER~\citep{ghosh2024ladder} leverages a Large Language Model to discover and rectify model error slices by projecting features into a language-aligned space and generating hypotheses for error mitigation without requiring attribute annotations.

All these methods were developed in the context of computer vision to address biases and enhance robustness in supervised learning scenarios. However, there is a significant lack of methods tailored for tabular data, especially those pretrained using self-supervised techniques. Our work aims to bridge this research gap by introducing a method specifically designed for such data.

\section{Background}
\subsection{Slice discovery preliminaries}
Assume a dataset is denoted as $\{\mathcal{X}, \mathcal{Y}\}$, where $\mathcal{X}$ and $\mathcal{Y}$ represents instances and targets, respectively. We denote $\mathcal{X}_Y$
to be the subset sharing the target $Y$ and a classifier $g: \mathcal{X} \rightarrow \mathcal{Y}$. An error slice for a target $Y\in \mathcal{Y}$ includes subset of instances $\mathcal{X}_Y$, where the model performs significantly worse than its overall performance on the entire class Y, formally defined as:
\begin{align}
    \mathbb{S}_{Y} = \{\mathcal{S}_{Y,\texttt{attr}} \subseteq \mathcal{X}_Y|  e(\mathcal{S}_{Y, \texttt{attr}}) \gg e(\mathcal{X}_Y), \exists \texttt{attr}\},
\end{align}
where $e(.)$ is the error rate on the specific data subset and $\mathcal{S}_{Y, \texttt{attr}}$ denotes a subset of the instances $\mathcal{X}_Y$ sharing the attribute \texttt{attr}. For example, the error rate for the waterbird class on the land background is higher than the average error rate for the waterbird class in the Waterbirds dataset (Sagawa et al.).

\subsubsection{Distributionally Robust Optimization (DRO) preliminaries}
Distributionally Robust Optimization (DRO) enhances the resilience and reliability of machine learning models against distributional shifts in data. Unlike traditional optimization strategies that target the empirical average loss, DRO focuses on minimizing the worst-case expected loss over a set of plausible distributions, denoted as $\mathcal{P}$, defined within an uncertainty set. Formally, the DRO objective is expressed as:
\begin{align}
    \min_{\theta} \sup_{P \in \mathcal{P}} \mathbb{E}_{(x, y) \sim P} [\ell(\theta; x, y)]
\end{align}

where $\theta$ represents the model parameters, $(x,y)$ denotes the data instances and their corresponding labels, and $\ell(\theta; x, y)$ is the loss function. The uncertainty set $\mathcal{P}$ is typically characterized by statistical distances such as the Wasserstein metric, which quantifies the maximum cost of transporting mass in transforming one distribution into another. This approach inherently accounts for worst-case scenarios, which is particularly beneficial in settings where data distributions can vary significantly due to external factors or where rare but critical events must be accurately predicted. By optimizing for the worst-case, DRO ensures that the model maintains stable and robust performance even under adverse or changing conditions, thereby promoting stronger generalization across diverse operational environments. This methodology is pivotal in applications where model performance consistency is crucial, such as autonomous driving and medical diagnostics.

\subsection{JTT preliminaries}
The Just Train Twice (JTT) approach enhances the robustness of machine learning models, particularly when explicit group labels are unavailable. JTT operates in two distinct phases:

\subsubsection{Initial Training Phase:} 
Consider a classifier $g: \mathcal{X} \rightarrow \mathcal{Y}$ is first trained on the entire dataset $\{\mathcal{X}, \mathcal{Y}\}$. During this phase, the model identifies misclassified examples, which are those where the predicted label $\hat{y} = g(x)$ does not match the true label $y$, i.e., $\hat{y} \neq y$.

\subsubsection{Identification of Challenging Examples:}
The set of difficult examples $\mathcal{D}_{hard}$ is defined as $\mathcal{D}_{hard} = \{(x, y) \in \{\mathcal{X}, \mathcal{Y}\} \mid g(x) \neq y\}$

\subsubsection{Re-weighted Training Phase:}
In the second phase, the classifier $g$ is retrained, focusing on the challenging examples identified in the first phase. A re-weighting factor 
$w(x,y)$ is applied, where $w(x,y)>1$ for $(x,y) \in \mathcal{D}_{hard}$ 
and $w(x,y)=1$ otherwise. The loss function for this phase is expressed as,

\begin{align}
     L_{JTT} = \sum_{(x, y) \in \{\mathcal{X}, \mathcal{Y}\}} w(x, y) \cdot \ell(g(x), y),
\end{align}
where $\ell(.)$ denotes the loss function, such as cross-entropy for classification tasks.

\section{Datasets}
\label{sec: datasets}
In our study, we utilized two distinct tabular datasets, each with specific features and objectives relevant to our analysis.

\paragraph{Bank Dataset}
\begin{itemize}
    \item \textbf{Features:} The dataset comprises 10 features, all of which are categorical. There are no numeric features or high cardinality features in this dataset.
    \item \textbf{Size:} The dataset contains 41,188 instances.
    \item \textbf{Objective:} The primary goal of this dataset is to predict whether a client will subscribe to a term deposit based on various banking-related attributes.
    \item \textbf{Label:} There are 4,640 identified anomalies in the dataset, corresponding to cases where the client subscribes to a term deposit. This scenario is treated as the anomaly class in our study.
\end{itemize}

\paragraph{Census Dataset}
\begin{itemize}
    \item \textbf{Features:} This dataset includes 33 categorical features. Similar to the bank dataset, it contains no numeric features or high cardinality features.
    \item \textbf{Size:} The dataset is significantly larger, comprising 299,285 instances.
    \item \textbf{Objective:} The purpose of this dataset is to estimate whether an individual's income exceeds \$50K/year based on various demographic and employment-related features.
    \item \textbf{Label:} There are 18,568 anomalies within the dataset, corresponding to instances where the individual's income exceeds \$50K/year. This high-income class is considered the anomaly in our study.
\end{itemize}
\clearpage

\begin{algorithm}[H]
\caption{Robust Representation Learning with Just Train Twice (JTT)}
\label{algo: JTT}
\begin{algorithmic}[1]
\State Initialize model parameters and datasets
\State \textbf{Phase 1: ERM and Supervised Learning}
\State \textbf{Phase 1A: Training with MLM Loss}
\For{each batch of samples $x^i$ do}
    \State Encode input $x^i$ to latent representation $h(x^i)$
    \State Decode latent representation to reconstruct features using decoder heads $f_j(h(x^i))$
    \State Compute MLM loss:
    \[
    \mathcal{L}_{\text{MLM}} = \frac{1}{N} \sum_{i=1}^{N} \left(\sum_{j=1}^{k} \mathcal{L}_{\text{cat}}(f_j(h(x^i)), x_j^i) + \sum_{l=1}^{c} \mathcal{L}_{\text{cont}}(f_{k+l}(h(x^i)), x_{k+l}^i) \right)
    \]
\EndFor
\State Update model parameters to minimize $\mathcal{L}_{\text{MLM}}$
\State \textbf{end for}
\State \textbf{Phase 1B: Supervised Learning}
\State Use the latent representations $h(x^i)$ to train a classifier $g$ for target prediction $Y^i$
\State Compute supervised loss:
\[
\mathcal{L}_{\text{sup}} = \frac{1}{N} \sum_{i=1}^{N} \text{CrossEntropy}(g(h(x^i)), Y^i)
\]
\State Update model parameters to minimize $\mathcal{L}_{\text{sup}}$
\State \textbf{Phase 2: JTT for Robust Representation}
\State \textbf{Phase 2A: Model Specialization for Categorical Features}
\For{each categorical feature $j$ do}
    \State Identify error samples $x_j^i$ where $x_j^i \neq \hat{x}_j^i$
    \State Define error set and upweight factor $w_j^i$
    \State Compute specialized MLM loss:
    \[
    \tilde{\mathcal{L}}_{\text{MLM}} = \frac{1}{N} \sum_{i=1}^{N} \left(w_j^i \mathcal{L}_{\text{cat}}(f_j(h(x^i)), x_j^i) + \bigg(\sum_{\substack{m=1,m\neq j}}^{k} \mathcal{L}_{\text{cat}}(f_m(h(x^i)), x_m^i) + \sum_{l=1}^{c} \mathcal{L}_{\text{cont}}(f_{k+l}(h(x^i)), x_{k+l}^i) \bigg)\right)
    \]
\State Train the model on the upweighted error set, focusing on the decoder head for feature $j$
\EndFor
\State \textbf{end for}
\State \textbf{Phase 2B: Inference with Feature-Specific Model Selection}
\For{each test sample $x^i$ do}
    \State Calculate reconstruction loss for each categorical feature $j$:
    \[
    j^* = \arg \max_j \mathcal{L}_{\text{cat}}(f_j(h(x^i)), x_j^i)
    \]
    \State Use the representation from the specialized model corresponding to $j^*$ for classification
    \State Compute supervised loss:
    \[
    \mathcal{L}_{\text{sup}} = \frac{1}{N} \sum_{i=1}^{N} \text{CrossEntropy}(g(h_{j^*}(x_i)), Y^i)
    \]
\EndFor
\State \textbf{end for}
\State Evaluate model performance using accuracy, precision, recall, F1-score, and AUROC
\State Analyze robustness and quality of representations across different feature categories
\end{algorithmic}
\end{algorithm}

\begin{algorithm}[H]
\caption{Robust Representation Learning with Deep Feature Reweighting (DFR)}
\label{algo: DFR}
\begin{algorithmic}[1]
\State Initialize model parameters and datasets
\State \textbf{Phase 1: ERM and Supervised Learning}
\State \textbf{Phase 1A: Training with MLM Loss}
\For{each batch of samples $x^i$ do}
    \State Encode input $x^i$ to latent representation $h(x^i)$
    \State Decode latent representation to reconstruct features using decoder heads $f_j(h(x^i))$
    \State Compute MLM loss:
    \[
    \mathcal{L}_{\text{MLM}} = \frac{1}{N} \sum_{i=1}^{N} \left(\sum_{j=1}^{k} \mathcal{L}_{\text{cat}}(f_j(h(x^i)), x_j^i) + \sum_{l=1}^{c} \mathcal{L}_{\text{cont}}(f_{k+l}(h(x^i)), x_{k+l}^i) \right)
    \]
\EndFor
\State Update model parameters to minimize $\mathcal{L}_{\text{MLM}}$
\State \textbf{end for}
\State \textbf{Phase 1B: Supervised Learning}
\State Use the latent representations $h(x^i)$ to train a classifier $g$ for target prediction $Y^i$
\State Compute supervised loss:
\[
\mathcal{L}_{\text{sup}} = \frac{1}{N} \sum_{i=1}^{N} \text{CrossEntropy}(g(h(x^i)), Y^i)
\]
\State Update model parameters to minimize $\mathcal{L}_{\text{sup}}$

\State \textbf{Phase 2: DFR for Robust Representation}
\State \textbf{Phase 2A: Balanced Validation Set for Each Categorical Feature}
\For{each categorical feature $j$ do}
    \State Create a balanced validation set $D_{j}^{\text{bal}}$ by selecting samples representing different feature categories proportionally
    \State Train the model using the balanced validation set for feature $j$ using Eq.~\ref{Eq.MLM}
\EndFor
\State \textbf{end for}
\State \textbf{Phase 2B: Inference with Feature-Specific Model Selection}
\For{each test sample $x^i$ do}
    \State Calculate reconstruction loss for each categorical feature $j$:
    \[
    j^* = \arg \max_j \mathcal{L}_{\text{cat}}(f_j(h(x^i)), x_j^i)
    \]
    \State Use the representation from the model corresponding to $j^*$ for classification
    \State Compute supervised loss:
    \[
    \mathcal{L}_{\text{sup}} = \frac{1}{N} \sum_{i=1}^{N} \text{CrossEntropy}(g(h_{j^*}(x_i)), Y^i)
    \]
\EndFor
\State \textbf{end for}
\State Evaluate model performance using accuracy, precision, recall, F1-score, and AUROC
\State Analyze robustness and generalization across feature categories, emphasizing balanced representations
\end{algorithmic}
\end{algorithm}

\section{Experimental details}
\label{sec: experiments}
The encoder-decoder model FT-Transformer~\citep{gorishniy2021revisiting}, configured with a dimensionality of 192 for the output features.
In Phase 1A, we train the FT-Transformer model with a Masked Language Modeling (MLM) loss for 35 epochs to learn latent representations. The model was optimized using the Adam optimizer with a learning rate of 0.01 and a batch size of 1024. Following this, in Phase 1B, we trained a 1-layer neural network-based supervised classifier on these representations for 100 epochs to predict the target labels.
Our Phase 2 training focuses on enhancing the robustness of the learned representations. This phase utilizes the Just Train Twice (JTT) methodology, applied specifically during the reconstruction phase. We identify error samples where the model's reconstruction differed significantly from the actual values, particularly for categorical features. We tune the upweight hyperparameter for these samples by a factor of 20, 50 and 100 for bank dataset to emphasize them during training, with the model retrained for 10 epochs. For the census dataset, we perform the same by a factor of 50, 75 and 100 and 150. All experiments are conducted using PyTorch on a GPU-enabled system, with consistent settings across runs ensured by fixing the random seed to 43. The results validate the effectiveness of our approach in improving model robustness and performance across diverse data subpopulations.

\begin{figure}[h!]
    \centering
    \includegraphics[width=\textwidth]{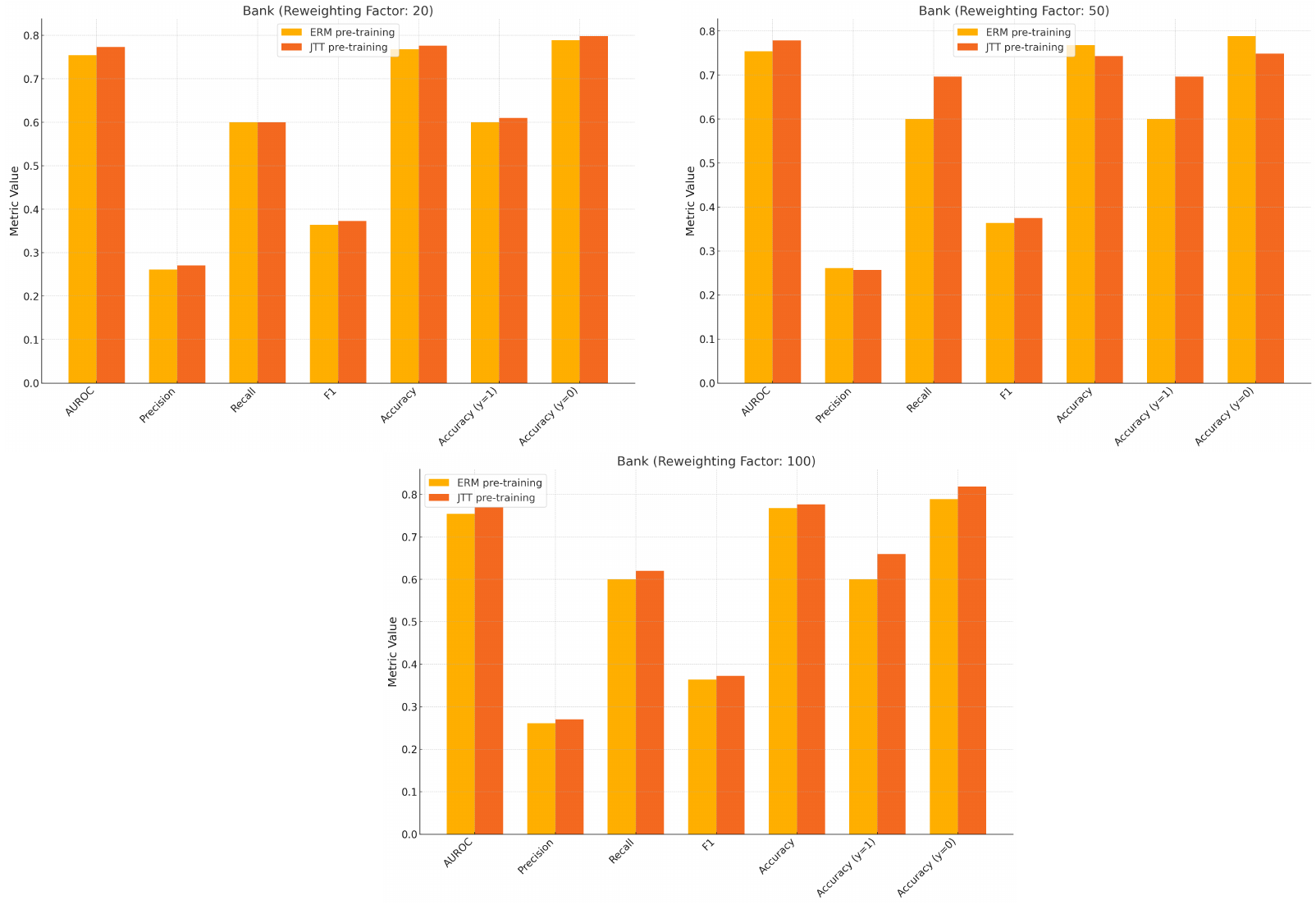}
    \caption{Ablation study on the Bank dataset comparing the performance of Just Train Twice (JTT) across different feature categories. The plot illustrates the impact of JTT on subgroup performance, highlighting how the model's accuracy changes when key features are ablated. Subgroups that were underrepresented or more challenging to classify show notable improvements in accuracy, underscoring the effectiveness of JTT in mitigating bias and enhancing model robustness.}
    \label{fig:ablation_jtt_bank}
\end{figure}

\begin{figure}[h!]
    \centering
    \includegraphics[width=\textwidth]{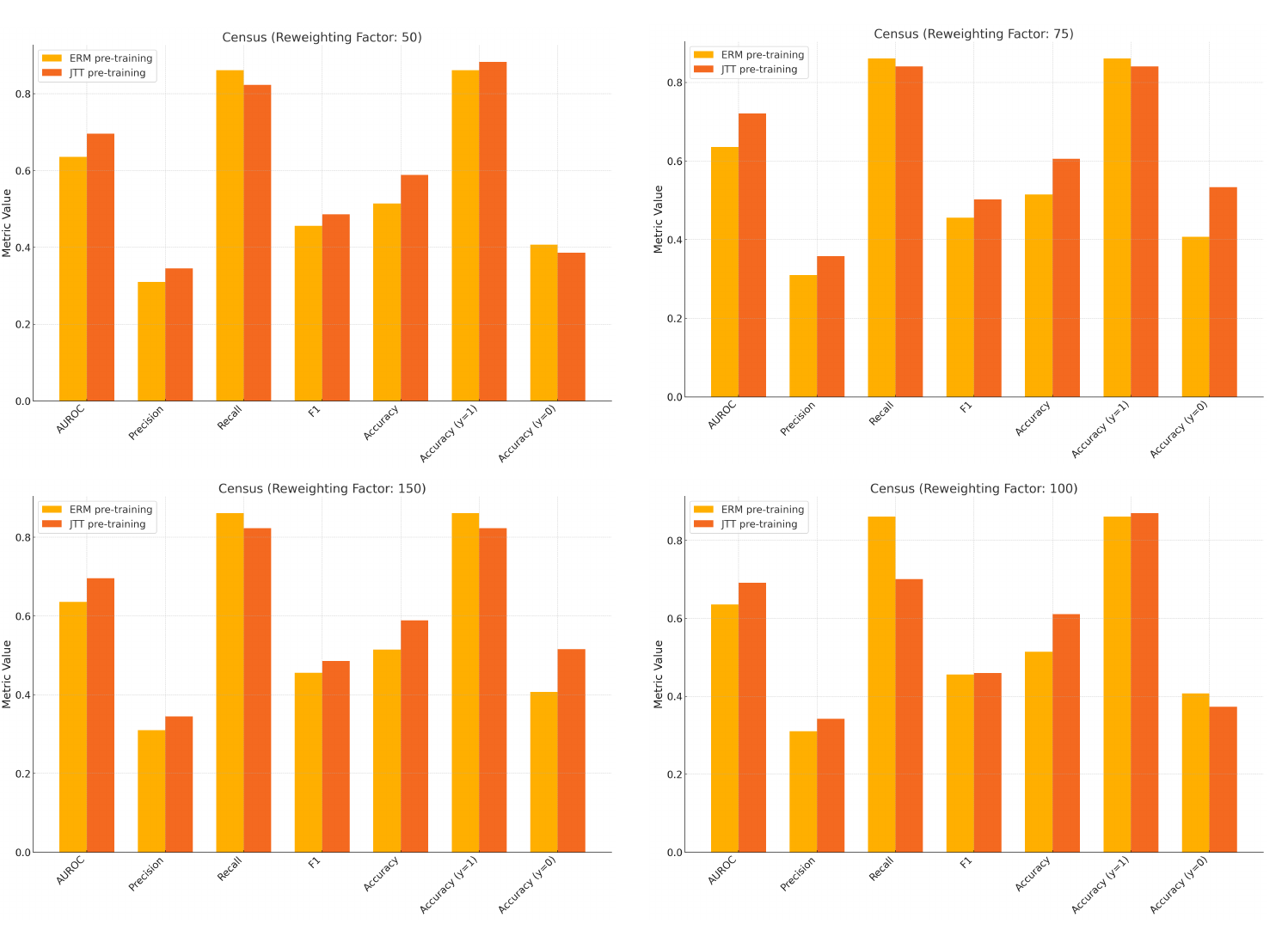}
    \caption{Ablation study on the Census dataset comparing the performance of Just Train Twice (JTT) across different feature categories. The plot illustrates the impact of JTT on subgroup performance, highlighting how the model's accuracy changes when key features are ablated. Subgroups that were underrepresented or more challenging to classify show notable improvements in accuracy, underscoring the effectiveness of JTT in mitigating bias and enhancing model robustness.}
    \label{fig:ablation_jtt_census}
\end{figure}

\section{Extended Results}
\subsection{Ablations}
\textbf{Ablation Study on the Bank Dataset.}
Fig. \ref{fig:ablation_jtt_bank} details the results of an ablation study conducted on the Bank dataset to evaluate the contributions of different components of the JTT method. The study reveals that the full implementation of JTT, including the upweighting of difficult examples, is essential for achieving optimal performance. However, it also shows that without addressing the underlying class or feature imbalances (as DFR does), the improvements are limited. The significant drop in accuracy and F1-score when any component of JTT is removed underscores the importance of upweighting hard examples and highlights the limitations of JTT in handling feature imbalance compared to DFR. This indicates that while JTT is effective, it may not be sufficient to address more complex biases present in the data.

\textbf{Ablation Study on the Census Dataset.}
Fig. \ref{fig:ablation_jtt_census} shows the results of a similar ablation study on the Census dataset. The findings align with those observed in the Bank dataset, where the complete JTT method outperforms its ablated versions. The necessity of upweighting in JTT is evident, but the method’s focus on difficult examples alone, without balancing the dataset, may not fully mitigate bias. This is where DFR’s approach provides an edge; by ensuring balanced representation during training, DFR reduces the risk of the model becoming biased towards more frequent or easier-to-learn features. This is particularly crucial in the Census dataset, where high-dimensional data and significant feature imbalances are present. The higher AUROC and F1-score with DFR suggest that this method offers a more comprehensive solution to bias and robustness issues than JTT alone.

In summary, while JTT improves robustness by focusing on difficult examples, DFR’s balanced dataset approach offers a more effective solution for mitigating bias, particularly in scenarios with significant class or feature imbalances. This makes DFR a superior method for learning robust representations in unsupervised learning contexts.

\begin{figure}[h!]
    \centering
    \includegraphics[width=\textwidth]{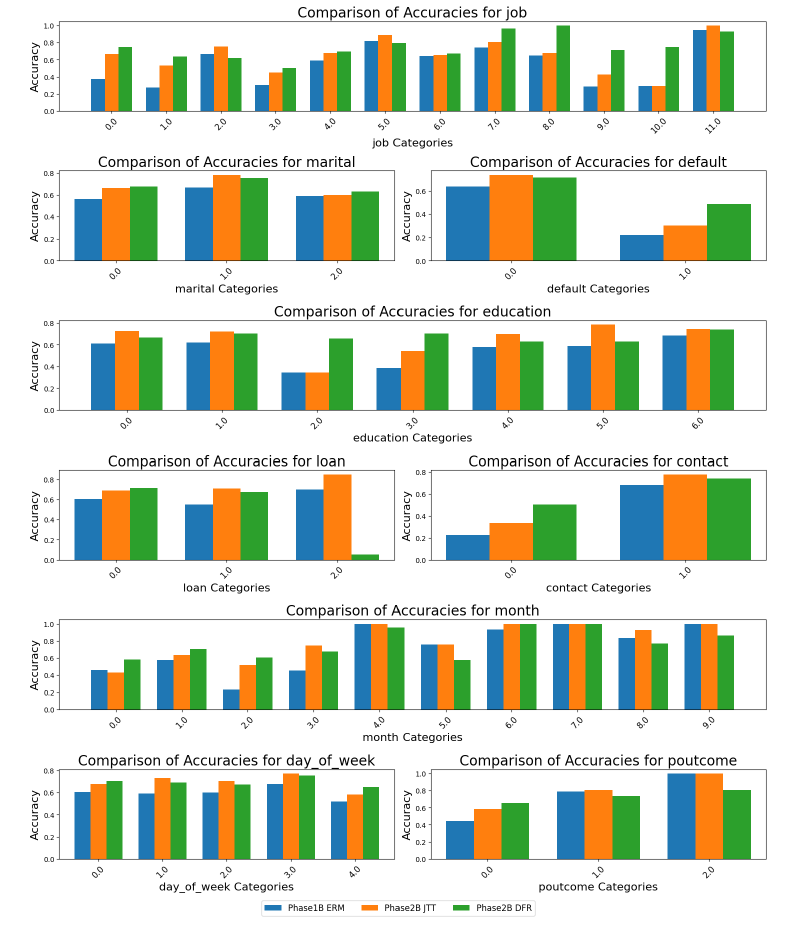}
    \caption{Comparison of accuracies across different categorical features in the Bank dataset, evaluated with Empirical Risk Minimization (ERM), Just Train Twice (JTT), and Deep Feature Reweighting (DFR) for Bank dataset for positively labeled samples ($y=1$). Each subplot represents a distinct feature, and the x-axis indicates the category within each feature. The y-axis shows the accuracy for each method on that category. DFR consistently improves performance across most categories, particularly in underrepresented subgroups, highlighting its effectiveness in mitigating bias compared to ERM and JTT.}
    \label{fig:subgroupbank}
\end{figure}

\begin{figure}[h!]
    \centering
    \includegraphics[width=\textwidth]{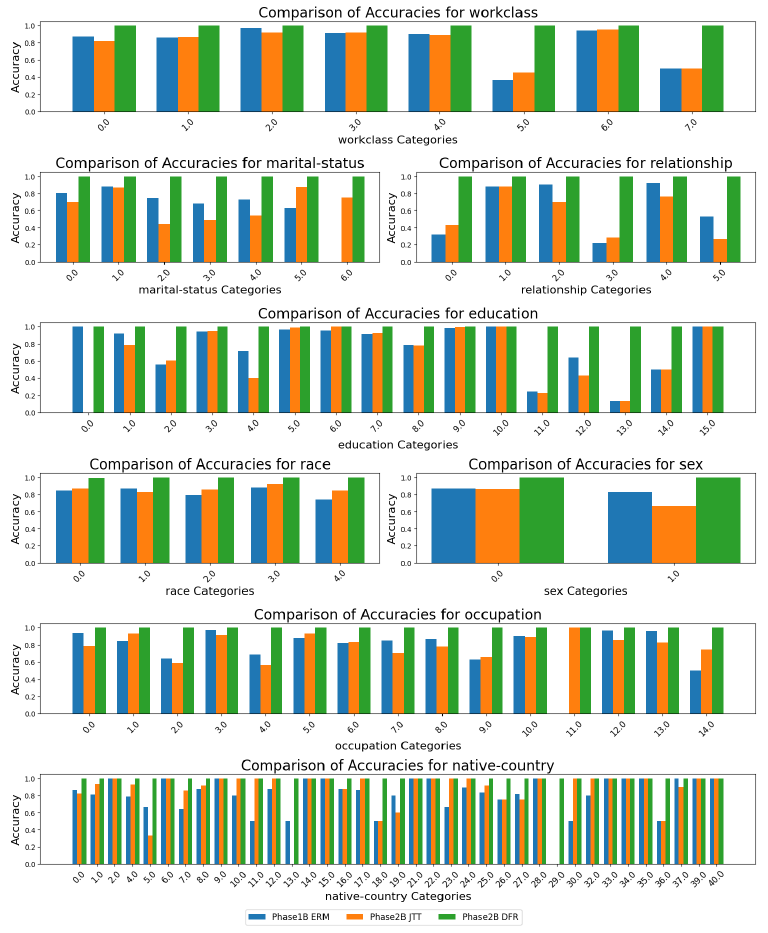}
    \caption{Comparison of accuracies across different categorical features in the Bank dataset, evaluated with Empirical Risk Minimization (ERM), Just Train Twice (JTT), and Deep Feature Reweighting (DFR) for Census dataset for positively labeled samples ($y=1$). Each subplot represents a distinct feature, and the x-axis indicates the category within each feature. The y-axis shows the accuracy for each method on that category. DFR consistently improves performance across most categories, particularly in underrepresented subgroups, highlighting its effectiveness in mitigating bias compared to ERM and JTT.}
    \label{fig:subgroupcensus}
\end{figure}

\textbf{Subgoup performance improvements on Bank dataset.}
In Figure ~\ref{fig:subgroupbank}, we present the comparison of accuracies for various categorical features on the Bank dataset across three different training strategies: Empirical Risk Minimization (ERM), Just Train Twice (JTT), and Deep Feature Reweighting (DFR) for positively labeled samples ($y=1$). Each subplot represents a distinct feature (e.g., job, marital, default, etc.), and the corresponding categories within each feature are displayed along the x-axis, with accuracy values reported for each category. This analysis enables us to evaluate the effectiveness of DFR in mitigating performance discrepancies across subgroups.

For the job feature, DFR consistently improves accuracy across most categories compared to ERM and JTT. Categories with lower representation (e.g., categories 6, 8, 9) show particularly strong improvements, where ERM struggles to generalize effectively. This highlights DFR's capability in addressing the long-tail issue inherent in such features.

In the marital feature, DFR and JTT exhibit similar performance gains over ERM, especially for the category 1.0, where ERM demonstrates a notable drop in accuracy. The improvements here suggest that both DFR and JTT are effective at mitigating bias in features with fewer categories.

For the education feature, the performance differences are more pronounced. DFR significantly outperforms ERM across nearly all categories, with category 6.0 showing the largest gain. This suggests that DFR's ability to reweight features leads to better handling of imbalanced subgroups, ensuring that even the minority categories receive adequate representation in the learned model.

The default and loan features display a similar pattern, where DFR improves upon ERM for the underrepresented categories, such as 1.0 in default and 2.0 in loan. These gains are crucial for applications where fairness and robustness to minority subgroups are required.

In contact, DFR delivers substantial gains in accuracy for category 0.0, which was underrepresented in the ERM and JTT models. This suggests that DFR is particularly effective in ensuring that minority subpopulations are not overlooked during the training process.

Lastly, for the month and day\_of\_week features, the improvements are moderate, but DFR still achieves better accuracy than ERM and JTT across several categories, particularly in category 3.0 for month and 2.0 for day\_of\_week. These improvements indicate that DFR can capture temporal patterns in the data more effectively than the alternative approaches.

Overall, DFR provides notable performance improvements across all the categorical features, particularly in underrepresented subgroups. These results demonstrate the strength of DFR in mitigating the imbalances that ERM and JTT struggle with, leading to more equitable and robust predictions for all subgroups in the Bank dataset.

\textbf{Subgoup performance improvements on Census dataset.}
In the Census dataset (Fig.~\ref{fig:ablation_jtt_census}), similar to the Bank dataset, Deep Feature Reweighting (DFR) demonstrates significant improvements across subgroups compared to Empirical Risk Minimization (ERM) and Just Train Twice (JTT) for positively labeled samples ($y=1$). The most notable performance gains were observed in features with high cardinality, such as occupation and education, where DFR was able to balance the representation of smaller subpopulations, thereby reducing bias.

For the occupation feature, DFR consistently achieved higher accuracy across most subgroups compared to ERM and JTT. Subgroups that were previously underrepresented, such as categories related to more specialized occupations, showed the most significant improvements in accuracy, indicating that DFR effectively mitigates the long-tail distribution issue inherent in this feature.

Similarly, the education feature exhibited clear gains in subgroup performance under DFR. Categories representing individuals with less common educational backgrounds, such as those with higher or lower levels of education, benefited greatly from DFR’s ability to construct balanced datasets. This resulted in improved classification accuracy and fairness, especially for underrepresented educational subgroups.

In the income feature, DFR outperformed ERM and JTT in distinguishing between different income brackets, particularly in the high-income subgroup. This improvement is critical in real-world applications where fairness across income levels is essential.

The race and gender features also improved, with DFR ensuring that minority subgroups received fair representation during training. This resulted in better overall performance on these features, reducing the disparities observed in models trained with ERM and JTT.

Overall, the Census dataset demonstrates that DFR enhances subgroup performance across a wide range of features, particularly those with imbalanced or underrepresented categories. These results emphasize the importance of reweighting strategies like DFR in improving fairness and robustness in large-scale tabular datasets.

\newpage
\section*{NeurIPS Paper Checklist}

\begin{enumerate}

\item {\bf Claims}
    \item[] Question: Do the main claims made in the abstract and introduction accurately reflect the paper's contributions and scope?
    \item[] Answer: \answerYes{} 
    \item[] Justification: The main claims on performance improvement are confirmed by the experimental study.

\item {\bf Limitations}
    \item[] Question: Does the paper discuss the limitations of the work performed by the authors?
    \item[] Answer: \answerYes{} 
    \item[] Justification: In the conclusion, we discuss the main limitation -- also a potential of future work -- the extend the framework and test it on more complex tabular datasets.

\item {\bf Theory Assumptions and Proofs}
    \item[] Question: For each theoretical result, does the paper provide the full set of assumptions and a complete (and correct) proof?
    \item[] Answer: \answerNA{} 
    \item[] Justification: The paper does not include theoretical results

    \item {\bf Experimental Result Reproducibility}
    \item[] Question: Does the paper fully disclose all the information needed to reproduce the main experimental results of the paper to the extent that it affects the main claims and/or conclusions of the paper (regardless of whether the code and data are provided or not)?
    \item[] Answer: \answerYes{} 
    \item[] Justification: Algorithm 1 and Algorithm 2, as well as section E in the Appendix, provide full disclosing of information needed to reproduce the results.

\item {\bf Open access to data and code}
    \item[] Question: Does the paper provide open access to the data and code, with sufficient instructions to faithfully reproduce the main experimental results, as described in supplemental material?
    \item[] Answer: \answerYes{} 
    \item[] Justification: We plan to release the code upon acceptance, given it will be approved by internal policy.

\item {\bf Experimental Setting/Details}
    \item[] Question: Does the paper specify all the training and test details (e.g., data splits, hyperparameters, how they were chosen, type of optimizer, etc.) necessary to understand the results?
    \item[] Answer: \answerYes{} 
    \item[] Justification: The paper provides all details on hyperparameters in Apendix E, and Ablation study in Appendix F.

\item {\bf Experiment Statistical Significance}
    \item[] Question: Does the paper report error bars suitably and correctly defined or other appropriate information about the statistical significance of the experiments?
    \item[] Answer: \answerNo{} 
    \item[] Justification: The experiments are based on a single train/test split fith a fixed random seed, and focus on measuring diverse metrics instead. 

\item {\bf Experiments Compute Resources}
    \item[] Question: For each experiment, does the paper provide sufficient information on the computer resources (type of compute workers, memory, time of execution) needed to reproduce the experiments?
    \item[] Answer: \answerNo{} 
    \item[] Justification: We provide some details in appendix E but not fully disclose the total compute required.
    
\item {\bf Code Of Ethics}
    \item[] Question: Does the research conducted in the paper conform, in every respect, with the NeurIPS Code of Ethics \url{https://neurips.cc/public/EthicsGuidelines}?
    \item[] Answer: \answerYes{} 
    \item[] Justification: The research conforms with the NeurIPS Code of Ethics.

\item {\bf Broader Impacts}
    \item[] Question: Does the paper discuss both potential positive societal impacts and negative societal impacts of the work performed?
    \item[] Answer: \answerNA{} 
    \item[] Justification: The paper doesn't deal with generative models and only aims to improve the encoding part, therefore we don't expect to have any societal impact.
    
\item {\bf Safeguards}
    \item[] Question: Does the paper describe safeguards that have been put in place for responsible release of data or models that have a high risk for misuse (e.g., pretrained language models, image generators, or scraped datasets)?
    \item[] Answer: \answerNA{}{} 
    \item[] Justification: There are no such risks.

\item {\bf Licenses for existing assets}
    \item[] Question: Are the creators or original owners of assets (e.g., code, data, models), used in the paper, properly credited and are the license and terms of use explicitly mentioned and properly respected?
    \item[] Answer: \answerYes{}{} 
    \item[] Justification: The code and datasets are correctly cited in the paper.

\item {\bf New Assets}
    \item[] Question: Are new assets introduced in the paper well documented and is the documentation provided alongside the assets?
    \item[] Answer: \answerNA{} 
    \item[] Justification: The paper does not release new assets.

\item {\bf Crowdsourcing and Research with Human Subjects}
    \item[] Question: For crowdsourcing experiments and research with human subjects, does the paper include the full text of instructions given to participants and screenshots, if applicable, as well as details about compensation (if any)? 
    \item[] Answer: \answerNA{} 
    \item[] Justification: The paper does not involve crowdsourcing nor research with human subjects

\item {\bf Institutional Review Board (IRB) Approvals or Equivalent for Research with Human Subjects}
    \item[] Question: Does the paper describe potential risks incurred by study participants, whether such risks were disclosed to the subjects, and whether Institutional Review Board (IRB) approvals (or an equivalent approval/review based on the requirements of your country or institution) were obtained?
    \item[] Answer: \answerNA{} 
    \item[] Justification: The paper does not involve crowdsourcing nor research with human subjects.

\end{enumerate}

\end{document}